\documentclass[9pt,twocolumn,twoside]{osajnl}

\journal{josaa} 

\setboolean{shortarticle}{false} 

\ifthenelse{\boolean{shortarticle}}{\colorlet{color2}{color2b}}{\colorlet{color2}{color2}} 

\def\psnr{{\mathrm{PSNR}}}

\def\mse{{\mathrm{MSE}}}
\def\dop{{\mathrm{DoP}}}
\def\aop{{\mathrm{AoP}}}
\def\x{\mathbf{x}}
\def\I{\mathbf{I}}

\def\pbm3d{\mathrm{PBM3D}}

\newcommand{\argmin}{\operatornamewithlimits{arg\,min}}

\title{Denoising Imaging Polarimetry by an Adapted BM3D Method}

\author[1,2,*]{Alexander B. Tibbs}
\author[2]{Ilse M. Daly}
\author[2]{Nicholas W. Roberts}
\author[1]{David R. Bull}

\affil[1]{Department of Electrical and Electronic Engineering, University of Bristol, BS8 1UB}
\affil[2]{School of Biological Sciences, University of Bristol, BS8 1TQ}

\affil[*]{Corresponding author: alex.tibbs@bristol.ac.uk}

\dates{Compiled \today}

\doi{\url{ }}

\begin{abstract}

Imaging polarimetry allows more information to be extracted from a scene than conventional intensity or colour imaging. However, a major challenge of imaging polarimetry is image degradation due to noise. This paper investigates the mitigation of noise through denoising algorithms and compares existing denoising algorithms with a new method, based on BM3D. This algorithm, PBM3D, gives visual quality superior to the state of the art across all images and noise standard deviations tested. We show that denoising polarization images using PBM3D allows the degree of polarization to be more accurately calculated by comparing it to spectroscopy methods.
\end{abstract}

\setboolean{displaycopyright}{false}

\begin{document}

\maketitle
\thispagestyle{fancy}

\ifthenelse{\boolean{shortarticle}}{\ifthenelse{\boolean{singlecolumn}}{\abscontentformatted}{\abscontent}}{}

\section{Introduction}
\label{sec:into}

The polarization of light describes how light waves propagate through space~\cite{hecht_optics_2002}. Although different forms of polarization can occur, such as circular polarization, in this paper we focus only on linear polarization, the form that is abundant in nature. Light is said to be completely linearly polarized (or polarized for the purposes of this paper) when all waves travelling along the same path through space are oscillating within the same plane. If however, there is no correlation between the orientation of the waves, the light is described as unpolarized. Polarized and unpolarized light are the limiting cases of partially polarized light, which can be considered to be a mixture of polarized and unpolarized light.

The polarization of light can be altered by the processes of scattering and reflection. As a form of visual information it provides a fitness benefit such that many animals~\cite{horvath_polarized_2004,horvath_polarized_2014} use polarization sensitivity for a variety of task-specific behaviours such as: navigation~\cite{wehner_significance_2006}, communication~\cite{how_out_2014} and contrast enhancement~\cite{how_target_2015}. Inspired by nature, there are now many devices that capture images containing information about the polarization of light~\cite{taylor_underwater_2002,york_bioinspired_2014}, known as imaging polarimeters or polarization cameras. These have been used in a growing number of applications~\cite{snik_overview_2014} including mine detection~\cite{de_jong_polarized_2007}, surveillance~\cite{lin_polarization_2004}, shape retrieval~\cite{miyazaki_shape_2007} and robot vision~\cite{shabayek_bio-inspired_2015}, as well as research in sensory biology~\cite{york_bioinspired_2014,britten_zebras_2016}.

A major challenge facing imaging polarimetry, addressed in this paper, is noise.
State of the art imaging polarimeters suffer from high levels of noise, and it will be shown that conventional image denoising algorithms are not well suited to polarization imagery.

Whilst a great deal of previous work has been done on denoising, very little has specifically been tailored to imaging polarimetry. Zhao et al.~\cite{zhao_new_2006} approached denoising imaging polarimetry by computing Stokes components from a noisy camera using spatially adaptive wavelet image fusion, whereas Faisan et al's~\cite{faisan_joint_2012} method is based on a modified version of the nonlocal means (NLM) algorithm~\cite{buades_review_2005}.

This paper compares the effectiveness of conventional denoising algorithms in the context of imaging polarimetry. A novel method termed PBM3D, adapted from an existing denoising algorithm, BM3D, is then introduced and will be shown to be superior to the state of the art. The use of appropriate test imagery will also be discussed.

\section{Imaging polarimetry}
\label{sec:polarimetry}
\subsection{Representing light polarization}
A polarizer is an optical filter which only transmits light of a given linear polarization. The angle between the transmitted light and the horizontal is known as the polarizer orientation. Let $I$ represent the total light intensity and $I_i$ represent the intensity of light which is transmitted through a polarizer orientated at $i$ degrees to the horizontal. The standard way of representing the linear polarization is by using Stokes parameters $(S_0,S_1,S_2)$~\cite{collett_field_2005}, which are defined as follows:
\begin{align}
S_0 &=I \\
S_1 &=I_0 - I_{90} \\
S_2 &=I_{45} - I_{135}.
\end{align}
Note that $I = I_0 + I_{90} = I_{45} + I_{135}$, so the above can be rewritten, using $I_0$, $I_{45}$ and $I_{90}$ as:
\begin{align}
S_0 &=I_0 + I_{90} \\
S_1 &=I_0 - I_{90} \\
S_2 &= - I_0 + 2I_{45} - I_{90}.
\end{align}

The degree of (linear) polarization ($\dop$) and the angle of polarization ($\aop$) are defined as:
\begin{align}
\dop &= \frac{\sqrt{S_1^2 + S_2^2}}{S_0} \label{eqn:dop}\\
\aop &= \frac{1}{2} \arctan\left(\frac{S_2}{S_1}\right).
\end{align}
The $\dop$ represents the proportion of light that is polarized, rather than being unpolarized, i.e. $\dop = 1$ means that the light is fully polarized, $\dop=0$ means unpolarized. The $\aop$ represents the average orientation of the oscillation of multiple waves of light, expressed as an angle from the horizontal.

\subsection{Imaging polarimeters}
Imaging polarimeters are devices which, in addition to measuring the intensity of light at each pixel in an array, also measure the polarization of light at each pixel location. There are many designs of imaging polarimeter, summarised in~\cite{tyo_review_2006}. The common feature they share is measuring the intensity of light which passes through polarizers of multiple orientations, $(I_{i_1},I_{i_2},\ldots,I_{i_n})$, possibly with additional measurements of circular polarization, at each pixel in an array. The measurements  for multiple orientations are taken either simultaneously or of a completely static scene. The Stokes parameters are then derived at each pixel. For the rest of this paper we will consider a polarimeter that measures $I_0$, $I_{45}$, and $I_{90}$, as this is the most common type~\cite{york_bioinspired_2014,taylor_underwater_2002}. Generalisations to imaging polarimeters which measure intensities at different angles are straightforward.

As this paper addresses polarization measurements across an array, the symbols $I_0$, $I_{45}$, $I_{90}$, $S_0$, $S_1$, $S_2$, $\dop$ and $\aop$ will henceforth refer to the array of values, rather than a single measurement. $I_0$, $I_{45}$ and $I_{90}$ are known as the camera components, and $S_0$, $S_1$ and $S_2$ as the Stokes components.

\subsection{Noise}
Noise affects most imaging systems and is especially challenging in polarimetry due to the complex sensor configuration involved with measuring the polarization. Each type of imaging polarimeter (see~\cite{tyo_review_2006} for a description of the different types) is affected by noise to a greater extend than conventional cameras are. `Division of focal plane' polarimeters, which use micro-optical arrays of polarization elements, suffer from imperfect fabrication and crosstalk between polarization elements. `Division of amplitude' polarimeters, which split the incident light into multiple optical paths, suffer from low SNR due to the splitting of the light. `Division of aperture' polarimeters, which use separate apertures for separate polarization components, suffer from distortions due to parallax. `Division of time' polarimeters require static scenes, and are incapable of recording video, so for many applications cannot be used. Also, in polarimetry, where $\dop$ and $\aop$ are often the quantities of interest, they are nonlinear functions of the camera and Stokes components and this has the effect of amplifying the noise degradation.

\begin{figure}
\includegraphics{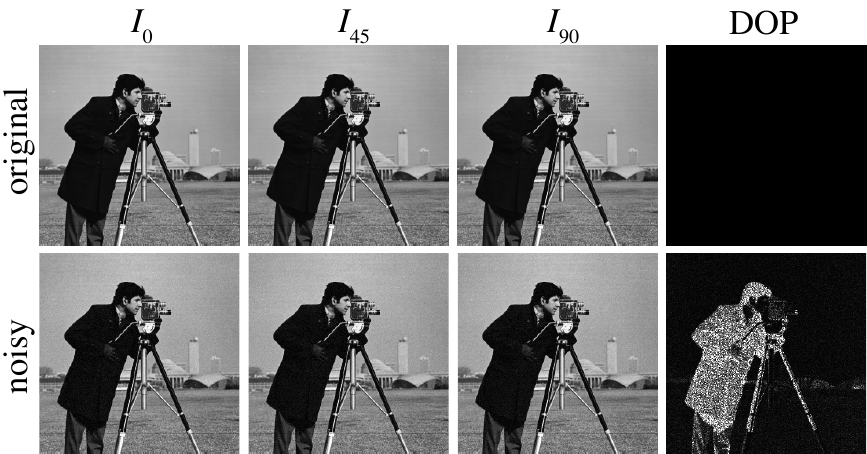}
\caption{Simulation of an unpolarized scene with and without noise ($\sigma = 0.02$). Black represents a value of 0, white of 1. The error is large for the noisy $\dop$ image.}
\label{fig:CameramanNoiseDop}
\end{figure}

To highlight the degradation of a $\dop$ image due to noise consider figure~\ref{fig:CameramanNoiseDop}. The top row shows the three camera components of an unpolarized scene (i.e. all three components are identical, and $\dop = 0$ everywhere). The original images with noise added are shown in the bottom row. Although there is only a small noticeable difference between the original and noisy camera components, the difference between the original and noisy $\dop$ images is severe. This indicates a large error, with 25\% of pixels exhibiting error greater than 10\%. The error is greater where the intensity of the camera components is smaller. To see why this is the case consider a noisy Stokes image $(S_0, S_1, S_2)$, where the measured values are normally distributed around the true Stokes parameters $(T_0,T_1,T_2)$. Let the true $\dop$ be given by $\delta_0 = (T_1^2 + T_2^2)^{1/2}/T_0$. The naive way to compute $\delta_0$ is to apply the $\dop$ formula to the measured Stokes parameters $\delta = (S_1^2 + S_2^2)^{1/2}/S_0$. But $\mathbb{E}(\delta) \ne \delta_0$ (where $\mathbb{E}$ is expected value) meaning that this is a biased estimator, so the calculated $\dop$ does not average to the correct result. This can be seen by the fact that if the true $\dop$, $\delta_0$, is zero, then any error in $S_0$ and $S_1$ results in $\delta > 0$. Denoising algorithms, including the one proposed in this paper, PBM3D, are thus essential for mitigating such degradations due to noise.

%


This paper considers only Gaussian noise, as this is typical of that which affects imaging polarimeters. A noisy camera component, $I_i$, is described as follows. Let $\Omega$ denote the image domain. For all $\x \in \Omega$ and $i \in \{0,45,90\}: I_i(\x) = I_i'(\x) + n(\x)$ where the noise, $n(\x)$, is a normally distributed zero-mean random variable with standard deviation $\sigma$, and $I_i'$ is the true camera component.

\section{State of the art}
\label{sec:State}

\subsection{Polarization denoising algorithms}

There are various methods for mitigating noise degradation in imaging polarimetry. For example, polarizer orientations can be chosen optimally for noise reduction~\cite{zallat_optimal_2006,xia_optimization_2015}, however this is not always possible due to constraints on the imaging system. Further reductions in noise can also be made through the use of denoising algorithms, which attempt to estimate the original image.

While there exists a vast literature on denoising algorithms in general, little is specifically targeted at denoising imaging polarimetry. Zhao et al.~\cite{zhao_new_2006} approached denoising imaging polarimetry by computing Stokes components from a noisy camera using spatially adaptive wavelet image fusion, based on~\cite{chang_spatially_2000}. A benefit of this algorithm is that the noisy camera components need not be registered prior to denoising. The algorithm of Faisan et al.~\cite{faisan_joint_2012} is based on a modified version of Buades et al.'s nonlocal means (NLM) algorithm~\cite{buades_review_2005}. The NLM algorithm is modified by reducing the contribution of outlier patches in the weighted average, and by taking into account the constraints arising from the Stokes components having to be mutually compatible. A disadvantage of this method is that it takes a long time to denoise a single image (550s for a $256 \times 256$ image which takes approximately 1s using our method, PBM3D. Both on an Intel Core i7, running at 3 GHz).

In this paper, our PBM3D algorithm will be compared to the above two algorithms. Faisan et al.~\cite{faisan_joint_2012} compared their denoising algorithm with earlier methods~\cite{sfikas_recovery_2011,valenzuela_joint_2009,zallat_polarimetric_2007} and demonstrated that their NLM based algorithm gives superior denoising performance. For this reason, comparison to these algorithms is not considered.

\subsection{Test imagery}
\label{sec:test}
In order to demonstrate the effectiveness of denoising algorithms, they must be evaluated using representative noisy test imagery. There are three types of test imagery used in the literature. The first is genuine polarization imagery captured by imaging polarimeters. This is ultimately what will be denoised, but it only allows for qualitative visual assessment as there is no ground truth, i.e. the original noise-free image is not available. This means that PSNR (peak signal to noise ratio) or other reference-based image quality metrics cannot be calculated. In order to have a ground truth for quantitative analysis, both the noisy and noise-free versions of the same image must be available.

The second type of test imagery comprises synthetic images with simulated noise. Synthetic images have the advantage of being completely controllable, so the effect of varying any parameter on denoising performance can be investigated. The disadvantage of using synthetic images is that they may not be fully representative of natural images, which can lead to algorithms appearing more or less effective than they would be with real images, especially if the synthetic test images are overly simplified. The simulated noise added may also not accurately reflect the properties of the real noise.

The third type of test imagery comprises noise-free polarization images, with simulated noise. Although it is not possible to produce noise-free images with a noisy imaging polarimeter, there are approaches which allow the noise level to be reduced to arbitrarily small levels; this is discussed in section~\ref{sec:Experiments}. The advantage of this type of test imagery is that the images are natural, so are likely, and can be chosen to be, similar to what will ultimately be denoised, which is dependent on the final application. Using simulated noise, however, does mean that the noise properties may be unrealistic. Of the imaging polarimetry denoising papers, only Valenzuela and Fessler~\cite{valenzuela_joint_2009} used real polarization images with simulated noise. Sfikas et al., Zallat \& Heinrich, and Faisan et al. \cite{sfikas_recovery_2011,zallat_polarimetric_2007,faisan_joint_2012} used simple synthetic images, consisting of simple geometric shapes with regions of uniform, or smoothly varying, Stokes components. Such images are easy to denoise using basic uniform or smoothly varying regions, and as such they don't test the algorithms' ability to denoise natural images. Faisan et al.~\cite{zhao_new_2006} used only real polarimetric images, and as such no PSNR values were given, making quantitative analysis difficult.

\section{Method}
\label{sec:Method}

Our approach to denoising polarization images is to adapt Dabov's BM3D algorithm~\cite{dabov_image_2007} for use with imaging polarimetry, a novel method which we call PBM3D.

BM3D was chosen primarily for its robustness and effectiveness. Sadreazami et al.~\cite{sadreazami_study_2016} recently compared the performance of a large number of state of the art denoising algorithms, using three test images and four values of $\sigma$, the noise standard deviation. They showed that no one denoising algorithm of those tested always gave the greatest denoised PSNR. However BM3D was always able to give denoised PSNR values close to the best performing algorithm, and in more than half the cases was in the top two. Another appealing aspect of BM3D is that extensions have been published for color images (CBM3D)~\cite{dabov_color_2007}, multispectral images (MSPCA-BM3D)~\cite{danielyan_denoising_2010}, volumetric data (BM4D)~\cite{maggioni_nonlocal_2013} and video (VBM4D)~\cite{maggioni_video_2012}. This extensibility shows the versatility of the core algorithm. Sadreazami et al. found that CBM3D was the best performing algorithm for color images with high noise levels.

\subsection{BM3D}

BM3D consists of two steps. In step 1 a basic estimate of the denoised image is produced, step 2 then refines the estimate produced in step 1 to give the final estimate. Steps 1 and 2 both consist of the same basic substeps, show in algorithm~\ref{alg:bm3d}.

\begin{algorithm}
\caption{BM3D - single step}
\label{alg:bm3d}
\begin{algorithmic}[1]
\For{each block (rectangular neighbourhood of pixels) in noisy image}
\State find similar blocks across the image
\Comment{for step 1 this is done using the noisy image, for step 2 the basic estimate}
\State stack similar blocks to form 3D group
\State apply 3D transform to obtain sparse representation
\State apply filter to denoise
\Comment{for step 1 the filtering is done using a hard thresholding operation, and for step 2 a Wiener filter is used}
\State invert transform
\EndFor
\For{each pixel}
\State estimate single denoised value from values of multiple overlapping blocks
\EndFor
\State \Return denoised image
\end{algorithmic}
\end{algorithm}
BM3D is described more fully in~\cite{dabov_image_2007}, and thorough analysis is provided in~\cite{lebrun_analysis_2012}.

\subsection{CBM3D}
\begin{figure*}
\includegraphics{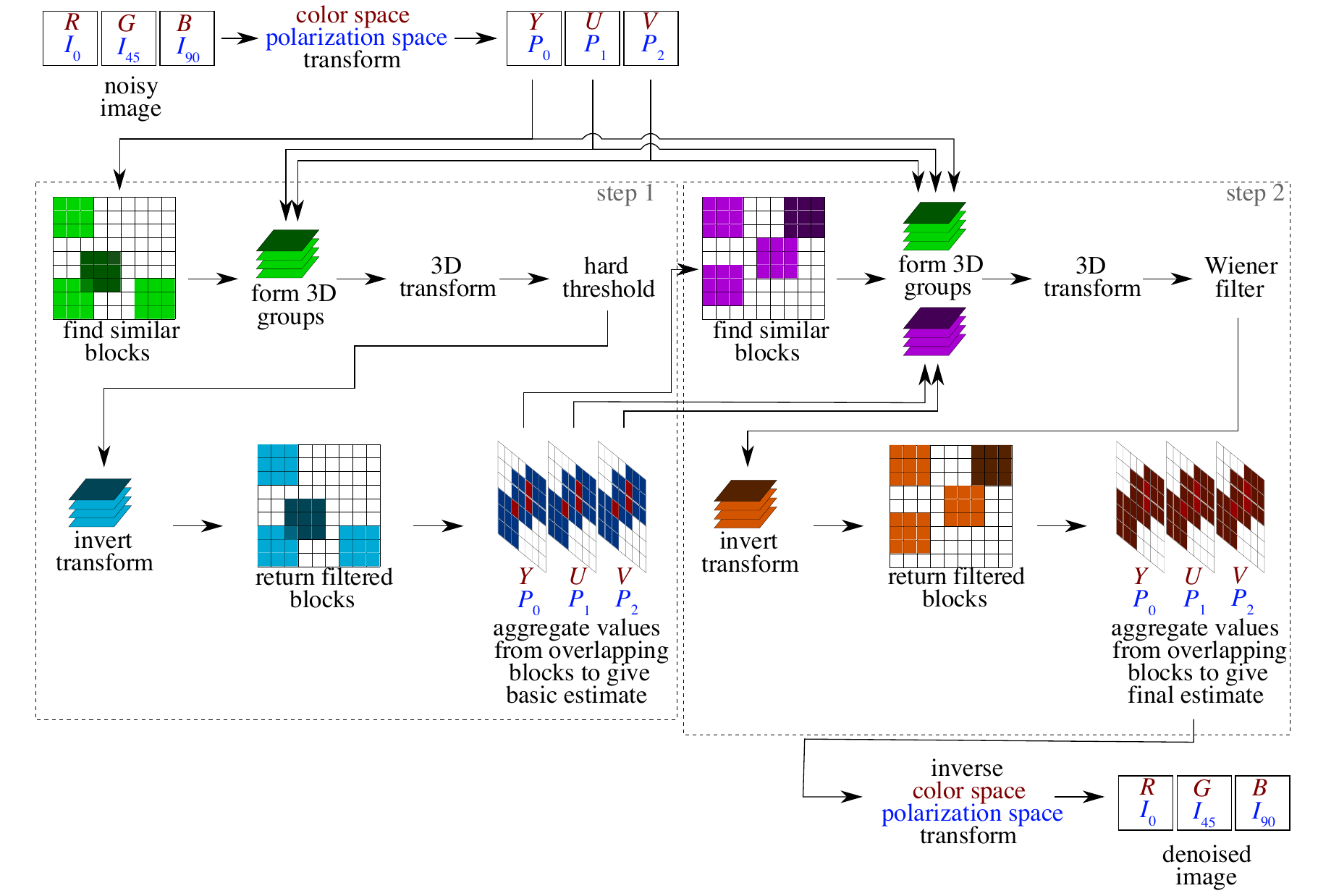}
\caption{Basic outline of the CBM3D/PBM3D denoising algorithm.}
\label{fig:bm3dDiagram}
\end{figure*}

CBM3D adapts BM3D for colour images~\cite{dabov_color_2007}. Figure~\ref{fig:bm3dDiagram} outlines the algorithm, which works by applying BM3D to the three channels of the image in the $YUV$ color space, also in two steps, but computing the groups only using the $Y$ channel. The details of CMB3D are as shown in algorithm~\ref{alg:cbm3d}.

\begin{algorithm}
\caption{CBM3D - single step}
\label{alg:cbm3d}
\begin{algorithmic}[1]
\State input noisy colour image
\State apply colour-space transform $(Y,U,V) \leftarrow T(R,G,B)$
\Comment{$YUV$ represents a chosen luminance-chrominance colour space}
\For{each block in channel $Y$ image}
\State find similar blocks across the image
\Comment{for step 1 this is done using the noisy image, for step 2 the basic estimate}
\State \label{alg:groupingCbm3d} stack similar blocks to form 3D group
\EndFor
\For{channels $U,V$}
\State stack blocks to form 3D groups using the same groups as formed in line~\ref{alg:groupingCbm3d}
\EndFor
\For{each channel $Y,U,V$}
\For{each group}
\State apply 3D transform to obtain sparse representation
\State apply filter to denoise
\Comment{for step 1 the filtering is done using a hard thresholding operation, and for step 2 a Wiener filter is used}
\State invert transform
\EndFor
\For{every pixel}
\State estimate single denoised value from values of multiple overlapping blocks
\EndFor
\EndFor
\State Apply inverse colour-space transform $(R,G,B) \leftarrow T^{-1}(Y,U,V)$
\State \Return denoised image
\end{algorithmic}
\end{algorithm}

Dabov et al \cite{dabov_color_2007} provide the following reason for why CBM3D performs better than applying BM3D separately to three colour channels:
\begin{itemize}
\item The SNR of the intensity channel, $Y$ is greater than the chrominance channels.
\item Most of the valuable information, such as edges, shades, objects and texture are contained in $Y$.
\item The information in $U$ and $V$ is tends to be low-frequency.
\item Isoluminant regions, with $U$ and $V$ varying are rare.
\end{itemize}

If BM3D is performed separately on colour channels, $U$ and $V$, the grouping suffers~\cite{dabov_color_2007} due to the lower SNR, and the denoising performance is worse as it is sensitive to the grouping.

\subsection{PBM3D}
\begin{figure}
\includegraphics{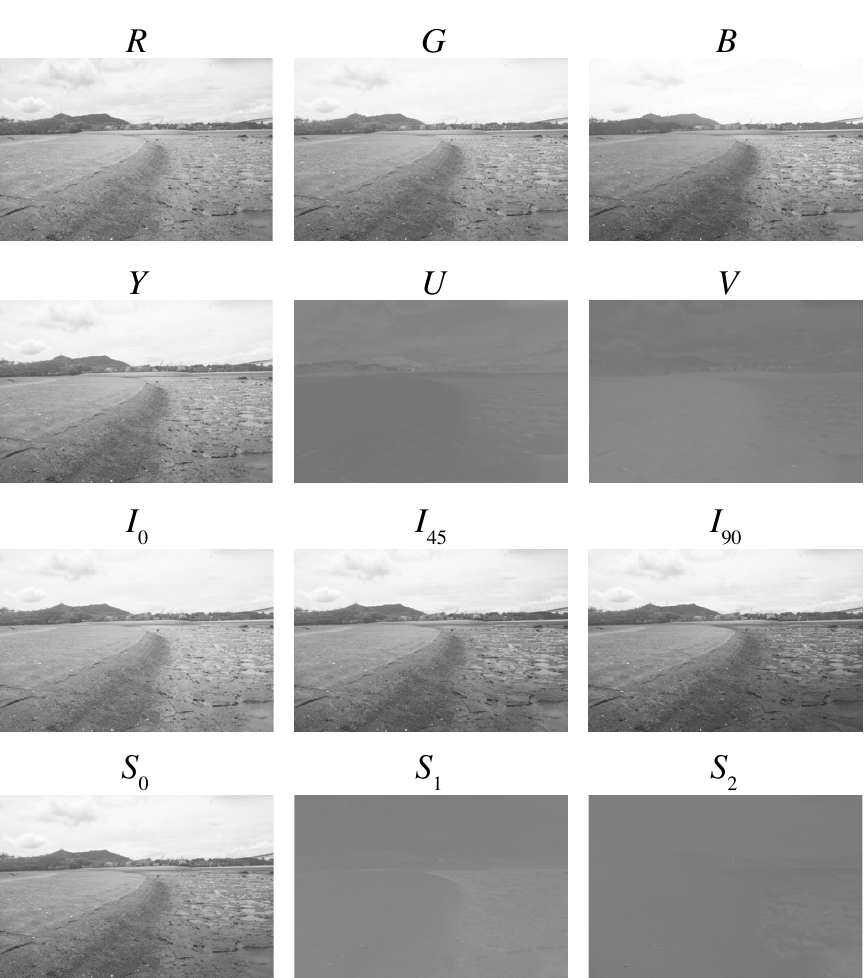}
\caption{Scene imaged in colour and in polarization. The top row shows the $RGB$ components of a colour image. The second row shows the same image in the $YUV$ color space. The third row shows the same scene as three polarization components $(I_0,I_{45},I_{90})$. The bottom row shows polarization image transformed into Stokes parameters. Photo credit: Martin How}
\label{fig:Mudflat}
\end{figure}

Colour imagery and polarization imagery are analogous in the way demonstrated in figure~\ref{fig:Mudflat}, where color and polarization components of the same scene are shown. The $R$, $G$, and $B$ images, like $I_0$, $I_{45}$, $I_{90}$ are visually similar to one another, each containing most of the information about the scene with small differences due to the colour in the case of $RGB$ and polarization in the case of $(I_0, I_{45}, I_{90})$. When $RGB$ is transformed into $YUV$, most of the information is contained in the luminance, $Y$. In much the same way, when the polarization components are transformed from $(I_0,I_{45},I_{90})$ into $(S_0,S_1,S_2)$ using the standard Stokes transformation, most of the information is contained in $S_0$. 

In order to optimise BM3D for polarization images, we propose taking CBM3D and replacing the $RGB$ to $YUV$ transformation with a transformation from the camera component image $(I_0,I_{45},I_{90})$ image to a chosen luminance-polarization space, denoted generally as $(P_0,P_1,P_2)$, as depicted in figure~\ref{fig:bm3dDiagram}. This is shown in algorithm~\ref{alg:pbm3d}.

\begin{algorithm}
\caption{PBM3D - single step}
\label{alg:pbm3d}
\begin{algorithmic}[1]
\State input noisy polarization image
\State apply polarization transform $(P_0,P_1,P_2) \leftarrow T(I_0,I_{45},I_{90})$
\Comment{$(P_0,P_1,P_2)$  represents a chosen luminance-polarization colour space}
\For{each block in channel $P_0$ image}
\State find similar blocks across the image
\Comment{for step 1 this is done using the noisy image, for step 2 the basic estimate}
\State \label{alg:groupingPbm3d} stack similar blocks to form 3D group
\EndFor
\For{channels $P_1,P_2$}
\State stack blocks to form 3D groups using the same groups as formed in line~\ref{alg:groupingPbm3d}
\EndFor
\For{each channel $(P_0,P_1,P_2)$}
\For{each group}
\State apply 3D transform to obtain sparse representation
\State apply filter to denoise
\Comment{for step 1 the filtering is done using a hard thresholding operation, and for step 2 a Wiener filter is used}
\State invert transform
\EndFor
\For{every pixel}
\State estimate single denoised value from values of multiple overlapping blocks
\EndFor
\EndFor
\State Apply inverse colour-space transform $(I_0,I_{45},I_{90}) \leftarrow T^{-1}(P_0,P_1,P_2)$
\State \Return denoised image
\end{algorithmic}
\end{algorithm}

The motivation for denoising polarimetry in this manner is the same as for colour images. The polarization parameters $(P_0,P_1,P_2)$ are analogous to the $(Y,U,V)$ components in the following ways:
\begin{itemize}
\item $P_0$, like $Y$, tends to have higher SNR than $P_1$ and $P_2$ in natural polarization images.
\item $P_0$, like $Y$, contains most of the valuable information.
\item Regions with constant $P_0$ and varying $P_1$ and $P_2$ are uncommon.
\end{itemize}

\subsection{Choice of polarization transformation}
\label{sec:Choice}

PBM3D relies on a linear transformation, represented by matrix $T$, to convert camera components into intensity-polarization components, the choice of $T$ has a large effect on denoising performance. Which matrix is optimal is dependent on the image statistics and the noise level, which are both dependent on the application. Here we describe how to estimate the optimal matrix, $T_{opt}$, given a set of noise-free model images, $D$, and a given noise standard deviation, $\sigma$.

Let $\I^i \in D$ be a noise free camera component image (e.g. $\mathbf{I} = (I_0,I_{45},I_{135})$), $\I^i{}'$ be $\I^i$ with Gaussian noise of standard deviation $\sigma$ added, $D'$ be the set of images $\I^i{}'$ and $\pbm3d_T$ represent the operation of applying PBM3D with transformation matrix $T$. Define $T_{opt}$ as follows:
\begin{equation}
\label{eqn:objective}
T_{opt} = \argmin_T \sum_{i} \mse(\I^i,\pbm3d_T (\I^i{}')),
\end{equation}
where $\mse$ is the mean square error. Note that $T$ is normalised such that for each row $\begin{pmatrix} a & b & c \end{pmatrix}$, $|a|+|b|+|c| = 1$.

Due to the large number of degrees of freedom of $T$, and the fact that the matrix elements can take any value in the range $[-1,1]$, it is not possible to perform an exhaustive search. Instead a Monte Carlo method, and a pattern search method can be used and are described here. Results from using both methods are shown in section~\ref{sec:Experiments}.

\subsubsection{Monte Carlo method}

This method has the advantages of being simple to implement, and not susceptible to convergence to local minima, but has the disadvantage of being slow to converge. It is shown in algorithm~\ref{alg:montecarlo}.

\begin{algorithm}
\caption{Monte Carlo method}
\label{alg:montecarlo}
\begin{algorithmic}[1]
\For{the desired number of rounds}
\State randomly choose a valid matrix $T$
\For{each image $\I^i \in D$}
\State denoise $\I^i$ using $T$
\State compute the mean MSE between every denoised image and its original
\EndFor
\EndFor
\State \Return $T_{opt}$, the matrix such that the mean MSE is smallest
\end{algorithmic}
\end{algorithm}

\subsubsection{Pattern search method}
This method has faster convergence, but can converge to local rather than global minima. It is shown in algorithm~\ref{alg:patternsearch}. Note that the intervals $\delta$ and $10\delta$ are both used to avoid converging to non-global minima.
\begin{algorithm}
\caption{Pattern search method}
\label{alg:patternsearch}
\begin{algorithmic}[1]
\State choose a starting matrix $T_0$ and small interval $\delta$
\State $i \leftarrow 0$
\Loop
\State find all perturbations of $T_i$ by $\delta$ which preserve the normalisation condition
\Comment{for each row $\begin{pmatrix} a & b & c \end{pmatrix}$, $|a|+|b|+|c| = 1$}
\State find all perturbations of $T_i$ by $10\delta$ which preserve the normalisation condition
\For{every perturbation, $P$, of $T_i$}
\For{every image, $\I' \in D'$}
\State denoise $\I'$ using $P$
\EndFor
\State $M_P \leftarrow \sum_{i}\mse(\I',\pbm3d_P(\I'))$
\EndFor
\State $T_{i+1} \leftarrow \argmin_P M_P$
\If{$T_{i+1} = T_i$}
\Return $T_{opt} \leftarrow T_i$
\EndIf
\State $i \leftarrow i+1$
\EndLoop
\end{algorithmic}
\end{algorithm}

\section{Experiments}
\label{sec:Experiments}

\subsection{Datasets}
\label{sec:datasets}
As noise-free polarization images cannot be produced using a noisy imaging polarimeter, we instead use a DSLR camera with a rotatable polarizer in front of the lens. This approach to producing imaging polarimetry is one of the earliest~\cite{wolff_polarization-based_1990} and has been used by various authors e.g.~\cite{how_target_2015,schechner_polarization-based_2003}.

For this approach to work, the camera sensor must have a linear response with respect to intensity, that is $I_{measured} = kI_{actual}$, where $I_{meaured}$ and $I_{actual}$ are the measured and actual light intensities, and $k$ is an arbitrary constant. The linearity can be verified using a fixed light source and a second rotating polarizer. As one polarizer is kept stationary, and the other is rotated, the intensity values measured at each pixel will produce a cosine squared curve if the sensor is linear, according to Malus' law~\cite{collett_field_2005}. The DSLR used to generate the images in this paper was a Nikon D70, whose sensors have a linear response.

The images are generated as follows:

\begin{algorithmic}[1]
\State All camera settings are set to manual for consistency between shots
\State To prevent inaccuracies due to compression the camera is set to take images in raw format
\State The camera is placed on a tripod or otherwise such that it is stationary
\State The polarizer is orientated to be parallel to the horizontal and an image is taken
\State The polarizer is rotated so that it is at $45^\circ$ to the horizontal and a second image is taken
\State The polarizer is rotated so that it is orientated vertically and the final image is taken
\end{algorithmic}

Given the superior SNR of modern DSLR cameras, this provides low noise polarization images. For arbitrarily low noise levels, multiple photos for a given polarizer angle are taken, registered and averaged. The main drawback of this method is that the light conditions and image subjects must be stationary, this method therefore cannot be used for many applications, but still allows noise free polarization images to be taken, so is invaluable for testing denoising algorithms.

\subsection{Optimal denoising matrix}
\label{sec:Optimal}

\begin{figure}
\includegraphics{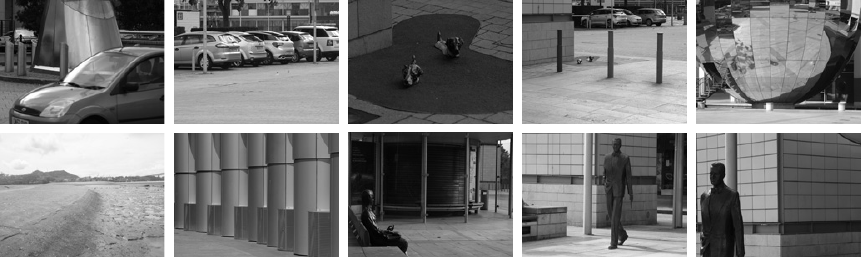}
\caption{$I_0$ of each image in the set.}
\label{fig:imageSet}
\end{figure}

The optimal matrix for a given application is dependent on the image statistics and the noise level. In order to test the matrix optimisation algorithms given in Section~\ref{sec:Method}, and with no particular application in mind we produced a set of 10 polarization images, using the method above, of various outdoor scenes. We added noise of several values of $\sigma$, the noise standard deviation (see tables~\ref{tab:MatrixOptResultsSubset}~\&~\ref{tab:OptMatrices}). The optimal matrices given in this section are therefore only optimal for this particular image set. However they provide a useful starting point and are likely to be close to optimal for applications where the images involve outdoor scenes. The choice of 10 images was arbitrary. Using a larger number of images would result in a more robust estimate of the optimal matrix. The $I_0$ component of each image is shown in figure~\ref{fig:imageSet}.

The natural choice of transform to gain an intensity-polarization representation of a polarization image from the camera components is to use a Stokes transformation, which, after normalization is given by:
\begin{equation}
T_{stokes} = \begin{pmatrix}
\frac{1}{2} & 0 & \frac{1}{2}  \\
\frac{1}{2} & 0 & -\frac{1}{2} \\
-\frac{1}{4} & \frac{1}{2} & -\frac{1}{4} \\
\end{pmatrix}.
\end{equation} However it was found during experiments that the opponent transform, $T_{opp}$ matrix of CBM3D~\cite{dabov_color_2007}, given below, almost always gives superior denoising performance to the Stokes transform.
\begin{equation}
T_{opp} = \begin{pmatrix}
\frac{1}{3} & \frac{1}{3} & \frac{1}{3}  \\
\frac{1}{2} & 0 & -\frac{1}{2} \\
\frac{1}{4} & -\frac{1}{2} & \frac{1}{4} \\
\end{pmatrix}
\end{equation} This is logical because taking the mean of the three camera components gives a greater SNR than taking the mean of only two components, and having greater SNR gives better grouping in the PBM3D algorithm, which is important as denoising performance is very sensitive to the quality of the grouping. The opponent matrix was therefore taken as the initial matrix, $T_0$ in the pattern search algorithm.

Both methods were applied to the model imagery at four $\sigma$ values. The Monte Carlo method was performed for 4000 rounds. The pattern search method was applied with $\delta = 0.01$. Table~\ref{tab:MatrixOptResultsSubset} shows the PSNR values for images denoised using the estimated optimal matrices. It can be seen that in almost every case the matrix found using the pattern search method results in the most effective denoising.

\begin{table*}
\begin{tabular}{@{}lccccccccccccccc@{}}
\toprule
& \multicolumn{5}{c}{street} & \multicolumn{5}{c}{dome} & \multicolumn{5}{c}{building} \\
\cmidrule(lr){2-6}
\cmidrule(lr){7-11}
\cmidrule(lr){12-16}
$\sigma$ & I & S & O & M & P & I & S & O & M & P & I & S & O & M & P \\
\midrule
0.01 & 32.6 & 32.6 & 32.6 & 46.0 & \textbf{46.2} & 44.8 & 45.7 & 46.5 & 46.6 & \textbf{46.8} & 40.9 & 41.2 & 41.5 & 46.6 & \textbf{47.0} \\
0.057 & 30.9 & 31.1 & 31.3 & 36.1 & \textbf{36.3} & 37.5 & 38.5 & 39.2 & 39.1 & \textbf{39.2} & 35.0 & 35.8 & 36.5 & 37.2 & \textbf{37.3} \\
0.1 & 28.7 & 29.2 & 29.4 & 31.7 & \textbf{31.8} & 32.9 & 33.1 & 33.5 & \textbf{33.8} & 33.6 & 31.0 & 32.5 & 33.3 & 33.4 & \textbf{33.6} \\
0.15 & 26.7 & 26.8 & 27.1 & 28.2 & \textbf{28.3} & 29.7 & 29.0 & 29.1 & 30.2 & \textbf{32.0} & 28.7 & 29.4 & 30.2 & 30.2 & \textbf{30.3} \\
\bottomrule \end{tabular}

\caption{PSNR values for images (street, dome, building) denoised using the following matrices: I: identity matrix; S: Stokes matrix; O: opponent matrix; M: Monte Carlo optimal; P: pattern search optimal. $\sigma$ is the noise standard deviation. Bold indicates maximum PSNR.}
\label{tab:MatrixOptResultsSubset}
\end{table*}

The pattern search method was then applied at 10 sigma values, giving an estimated optimal matrix for each (table ~\ref{tab:OptMatrices}).

\begin{table}
\begin{tabular}{@{}lc@{}}
\toprule
$\sigma$ & optimal matrix \\
\midrule
0.01 & $\begin{pmatrix} 0.323 & 0.363 & 0.313 \\ 0.500 & -0.210 & -0.290 \\ 0.150 & -0.500 & 0.350 \\  \end{pmatrix} $  \\
0.026 & $\begin{pmatrix} 0.323 & 0.363 & 0.313 \\ 0.500 & -0.210 & -0.290 \\ 0.150 & -0.500 & 0.350 \\  \end{pmatrix} $  \\
0.041 & $\begin{pmatrix} 0.323 & 0.363 & 0.313 \\ 0.500 & -0.230 & -0.270 \\ 0.160 & -0.500 & 0.340 \\  \end{pmatrix} $  \\
0.057 & $\begin{pmatrix} 0.323 & 0.363 & 0.313 \\ 0.500 & -0.210 & -0.290 \\ 0.150 & -0.500 & 0.350 \\  \end{pmatrix} $  \\
0.072 & $\begin{pmatrix} 0.323 & 0.363 & 0.313 \\ 0.510 & -0.010 & -0.480 \\ 0.250 & -0.510 & 0.240 \\  \end{pmatrix} $  \\
0.088 & $\begin{pmatrix} 0.323 & 0.363 & 0.313 \\ 0.300 & 0.210 & -0.490 \\ 0.240 & -0.520 & 0.240 \\  \end{pmatrix} $  \\
0.1 & $\begin{pmatrix} 0.323 & 0.373 & 0.303 \\ 0.420 & 0.080 & -0.500 \\ 0.250 & -0.510 & 0.240 \\  \end{pmatrix} $  \\
0.12 & $\begin{pmatrix} 0.343 & 0.353 & 0.303 \\ 0.480 & -0.120 & -0.400 \\ 0.130 & -0.520 & 0.350 \\  \end{pmatrix} $  \\
0.13 & $\begin{pmatrix} 0.333 & 0.333 & 0.333 \\ 0.480 & -0.230 & -0.290 \\ 0.040 & -0.530 & 0.430 \\  \end{pmatrix} $  \\
0.15 & $\begin{pmatrix} 0.343 & 0.353 & 0.303 \\ 0.480 & -0.120 & -0.400 \\ 0.130 & -0.520 & 0.350 \\  \end{pmatrix} $  \\
\bottomrule \end{tabular}
\caption{The optimal matrices computed using the pattern search method for 10 values of $\sigma$, the noise standard deviation.}
\label{tab:OptMatrices}
\end{table}

The pattern search method was also applied to an image set containing all 10 images, each with noise added of 10 different $\sigma$ values. The following matrix was found to be optimal on average across all $\sigma$ values:
\begin{equation}
T_{opt} = \begin{pmatrix}
0.3133  &  0.3833  &  0.3033\\
0.4800  &  0.0300  & -0.5100\\
0.2600  & -0.5200  &  0.2200\\
\end{pmatrix}.
\end{equation}

\subsection{Comparison of denoising algorithms}
The performance of PBM3D with a variety of images (different to those used for the matrix optimization) and noise levels was compared to the performance of several other denoising algorithms for polarization:
\begin{itemize}
\item BM3D: Standard BM3D for grayscale images applied individually to each camera component $(I_0,I_{45},I_{90})$
\item BM3D Stokes: Standard BM3D applied individually to each Stokes component $(S_0,S_1,S_2)$, found by transforming the camera components.
\item Zhao: Zhao et al's method~\cite{zhao_new_2006}
\item Faisan: Faisan et al's method~\cite{faisan_joint_2012}
\end{itemize}
In order to quantitatively compare the denoising performance PSNR was computed for each denoised image.

For Stokes image $(S_0,S_1,S_2)$ with ground truth given by $(S'_0,S'_1,S'_2)$, with $S_0(\x) \in [0,1]$, $S_1(\x) \in [-1,1]$, $S_2(\x) \in [-1,1]$ and $\x \in \Omega$, where $\Omega$ is the image domain, PSNR is given by 
\begin{equation}
\label{eqn:psnr}
\psnr = 10 \log_{10} \left( \frac{1}{\mse} \right),
\end{equation}
where
\begin{align}
\label{eqn:mse}
\mse =& \frac{1}{3MN} \sum_{\x \in \Omega} \left( \left(S_0(\x)-S_0'(\x)\right)^2 \right. \nonumber \\
& \left. + \frac{1}{2}\left(S_1(\x)-S_1'(\x)\right)^2 + \frac{1}{2}\left(S_2(\x)-S_2'(\x)\right)^2\right).
\end{align}

\begin{figure*}
\includegraphics{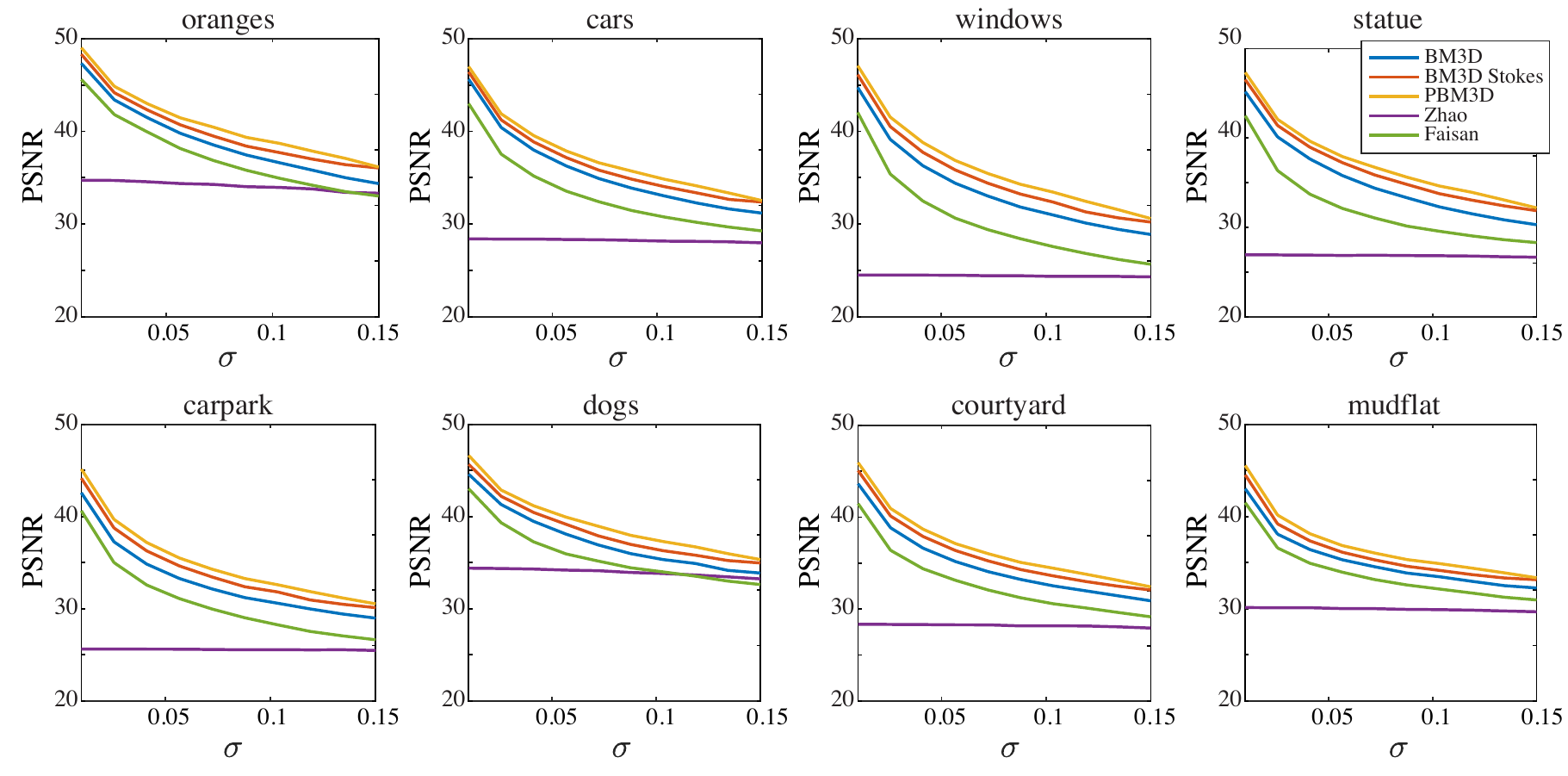}
\caption{PSNR for denoised images as a function of $\sigma$, the standard deviation of noise. Above each plot is the name of the image denoised, line colours correspond to different denoising algorithms. For the top row the PSNR values are shown in table~\ref{tab:BasicPbm3dMeanTab}. It can be seen that for all images and all values of $\sigma$ PBM3D produces images with the greatest PSNR.}
\label{fig:PsnrPlots}
\end{figure*}

\begin{figure}
\includegraphics{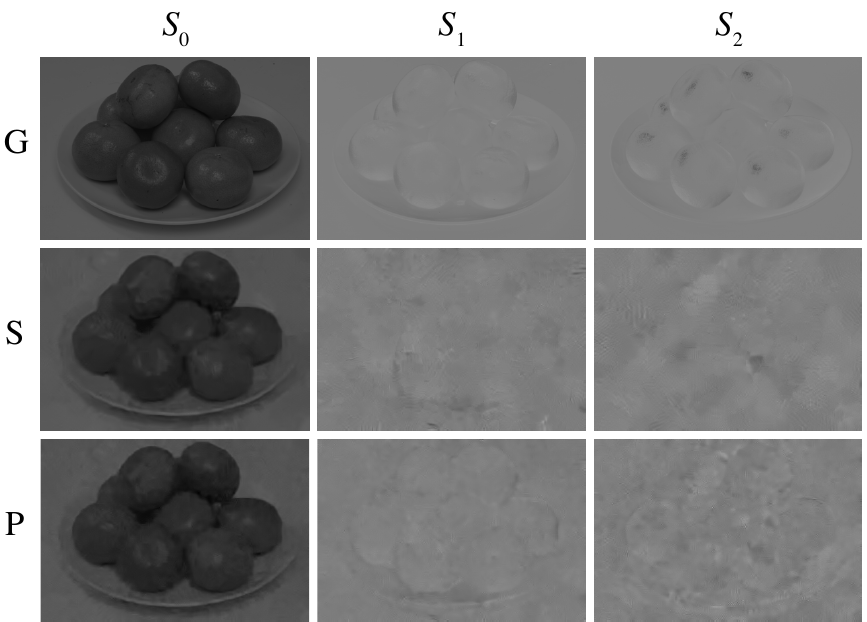}
\caption{`oranges' image with noise of standard deviation $\sigma = 0.15$ (a high noise level), denoised using BM3D Stokes (S) and PBM3D (P) (G: ground truth). For both BM3D Stokes and PBM3D, the $S_0$ images are visually similar to the ground truth. For both methods however, the $S_1$ images are notably different, and the $S_2$ images are almost unrecognisable.}
\label{fig:NoisyCloseUp}
\end{figure}

Table~\ref{tab:BasicPbm3dMeanTab} shows the PSNR values for four images (`oranges', `cars', `windows', `statue'). The same data, along with that for four other images is plotted in figure~\ref{fig:PsnrPlots}. It can be seen that PBM3D always provided the greatest denoising performance. Every image denoised using PBM3D at every noise level had a greater PSNR than images denoised using all other methods. The second best performing method in every case was BM3D Stokes, with PBM3D denoising images with a greater PSNR of 0.84dB on average. The difference in PSNR between images denoised using PBM3D and BM3D Stokes was greatest for the intermediate noise levels. The smaller noise levels exhibited less of a difference, and the PSNR values in the higher noise values became closer as noise was increased.The convergence of the PSNR values in the higher noise levels can be explained by the fact that the $S_1$ and $S_2$ components of the images become so noisy that there they bear little resemblance to the ground truth, as can be seen in figure~\ref{fig:NoisyCloseUp}. Zhao's method performed poorly at all noise levels, it provided a smaller PSNR at higher noise levels than the other methods at higher noise levels. Faisan's method had worse performance than all of the BM3D-based methods, at all noise levels (images denoised using Faisan had a PSNR 4.50dB smaller on average than those denoised using PBM3D), but performed significantly better than Zhao's method.

\begin{table*}
\setlength\tabcolsep{4.5pt}
\begin{tabular}{@{}lcccccccccccccccccccc@{}}
\toprule
& \multicolumn{5}{c}{oranges} & \multicolumn{5}{c}{cars} & \multicolumn{5}{c}{windows} & \multicolumn{5}{c}{statue} \\
\cmidrule(lr){2-6}
\cmidrule(lr){7-11}
\cmidrule(lr){12-16}
\cmidrule(lr){17-21}
$\sigma$ & B & S & P & Z & F & B & S & P & Z & F & B & S & P & Z & F & B & S & P & Z & F \\
\midrule
0.010 & 47.3 & 48.3 & \textbf{49.0} & 34.7 & 45.6 & 45.6 & 46.4 & \textbf{47.0} & 28.4 & 43.0 & 44.8 & 46.1 & \textbf{47.1} & 24.5 & 42.1 & 45.2 & 46.5 & \textbf{47.3} & 26.9 & 42.5 \\
0.026 & 43.4 & 44.2 & \textbf{44.9} & 34.7 & 41.8 & 40.4 & 41.2 & \textbf{41.9} & 28.4 & 37.5 & 39.1 & 40.5 & \textbf{41.5} & 24.5 & 35.4 & 40.1 & 41.4 & \textbf{42.1} & 26.9 & 36.4 \\
0.041 & 41.5 & 42.3 & \textbf{43.0} & 34.6 & 39.9 & 38.0 & 38.9 & \textbf{39.5} & 28.4 & 35.2 & 36.3 & 37.7 & \textbf{38.8} & 24.5 & 32.5 & 37.7 & 39.0 & \textbf{39.6} & 26.9 & 33.7 \\
0.057 & 39.8 & 40.7 & \textbf{41.5} & 34.4 & 38.1 & 36.3 & 37.2 & \textbf{37.9} & 28.3 & 33.5 & 34.4 & 35.8 & \textbf{36.9} & 24.5 & 30.6 & 35.8 & 37.2 & \textbf{37.9} & 26.9 & 32.1 \\
0.072 & 38.5 & 39.5 & \textbf{40.5} & 34.3 & 36.9 & 34.9 & 35.8 & \textbf{36.6} & 28.3 & 32.4 & 33.0 & 34.4 & \textbf{35.5} & 24.4 & 29.4 & 34.4 & 35.9 & \textbf{36.7} & 26.9 & 31.1 \\
0.088 & 37.4 & 38.4 & \textbf{39.4} & 34.0 & 35.8 & 33.9 & 34.8 & \textbf{35.7} & 28.2 & 31.5 & 31.8 & 33.2 & \textbf{34.3} & 24.4 & 28.4 & 33.3 & 34.8 & \textbf{35.6} & 26.8 & 30.1 \\
0.103 & 36.6 & 37.7 & \textbf{38.7} & 34.0 & 34.9 & 33.0 & 34.0 & \textbf{34.8} & 28.2 & 30.8 & 31.0 & 32.4 & \textbf{33.4} & 24.3 & 27.6 & 32.3 & 33.8 & \textbf{34.6} & 26.8 & 29.6 \\
0.119 & 35.8 & 37.0 & \textbf{37.9} & 33.8 & 34.2 & 32.3 & 33.4 & \textbf{34.1} & 28.1 & 30.2 & 30.1 & 31.3 & \textbf{32.5} & 24.4 & 26.8 & 31.5 & 33.1 & \textbf{33.9} & 26.8 & 29.0 \\
0.134 & 35.0 & 36.4 & \textbf{37.1} & 33.4 & 33.5 & 31.6 & 32.7 & \textbf{33.3} & 28.1 & 29.7 & 29.4 & 30.7 & \textbf{31.6} & 24.3 & 26.2 & 30.8 & 32.4 & \textbf{33.1} & 26.7 & 28.6 \\
0.150 & 34.4 & 36.1 & \textbf{36.2} & 33.3 & 33.0 & 31.2 & 32.4 & \textbf{32.5} & 28.0 & 29.3 & 28.9 & 30.2 & \textbf{30.6} & 24.3 & 25.7 & 30.3 & 31.9 & \textbf{32.2} & 26.7 & 28.3 \\
\bottomrule \end{tabular}
\caption{PSNR for denoising of four images (`oranges', `cars', `windows', `statue') using several methods (B: BM3D; S: BM3D Stokes; P: PBM3D; Z: Zhao; F: Faisan) and several values of $\sigma$, the standard deviation of the noise, added. Bold indicates greatest PSNR. It can be seen that PBM3D is always the best performing method. The pattern continues with other images, the results (including those shown here) are plotted in figure~\ref{fig:PsnrPlots}.}
\label{tab:BasicPbm3dMeanTab}
\end{table*}

\begin{figure}
\includegraphics{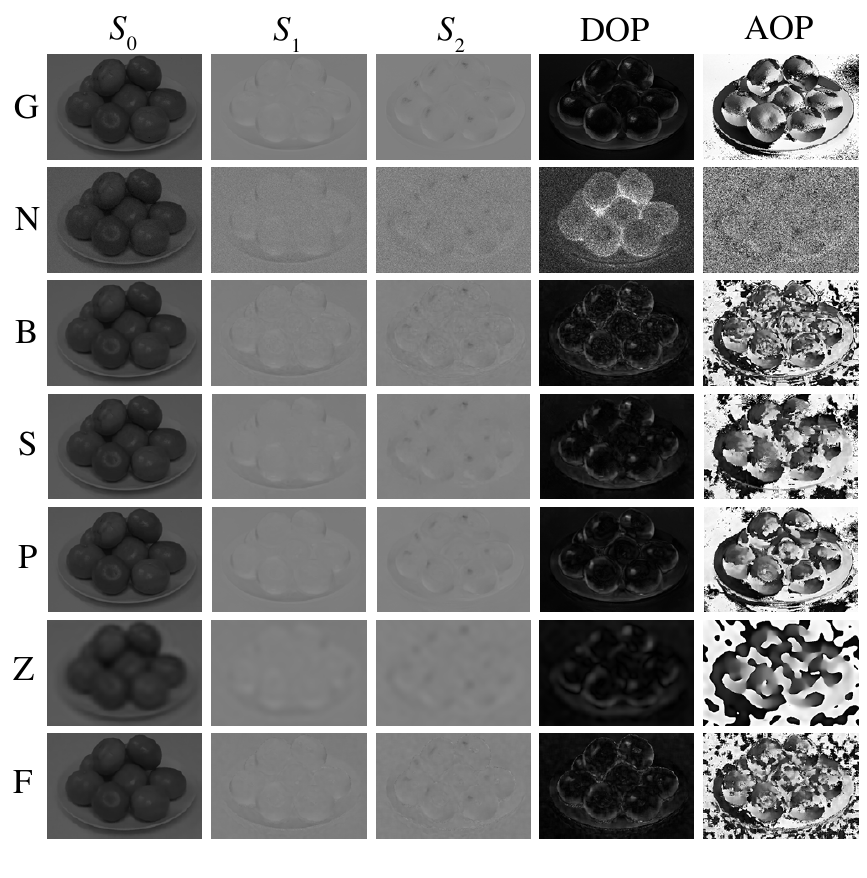}
\caption{Polarization components of `oranges' image after application of several denoising methods. G: ground truth; N: noisy; B: BM3D; S: BM3D Stokes; P: PBM3D; Z: Zhao; F: Faisan. Noise standard deviation, $\sigma = 0.026$. Note that the $\dop$ images have been scaled such that black represents $\dop=0$ and white represents $\dop=0.5$.}
\label{fig:simulatedImage1}
\end{figure}

\begin{figure}
\includegraphics{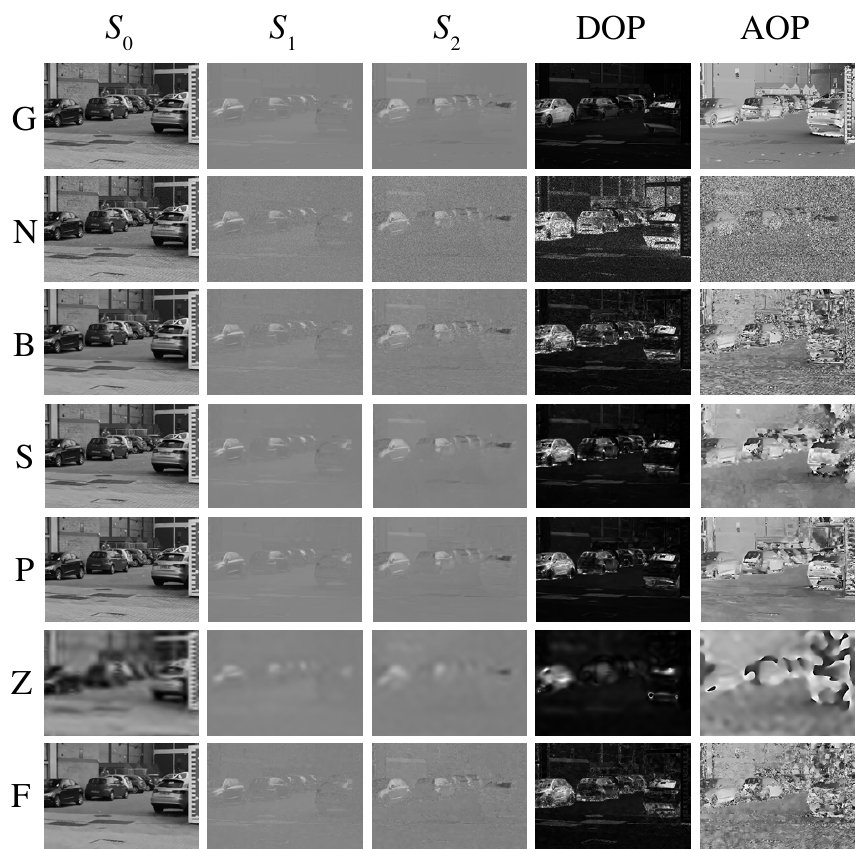}
\caption{Polarization components of `cars' image after application of several denoising methods. G: ground truth; N: noisy; B: BM3D; S: BM3D Stokes; P: PBM3D; Z: Zhao; F: Faisan. Noise standard deviation, $\sigma = 0.026$}
\label{fig:simulatedImage4}
\end{figure}

\begin{figure}
\includegraphics{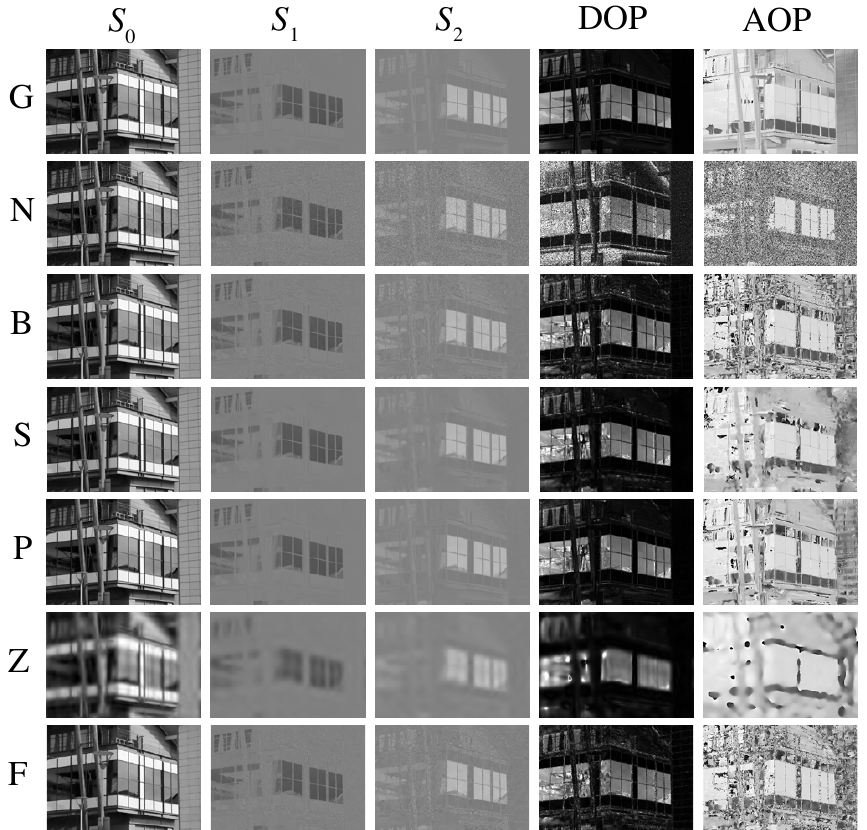}
\caption{Polarization components of `window' image after application of several denoising methods. G: ground truth; N: noisy; B: BM3D; S: BM3D Stokes; P: PBM3D; Z: Zhao; F: Faisan. Noise standard deviation, $\sigma = 0.026$}
\label{fig:simulatedImage8}
\end{figure}

\begin{figure}
\includegraphics{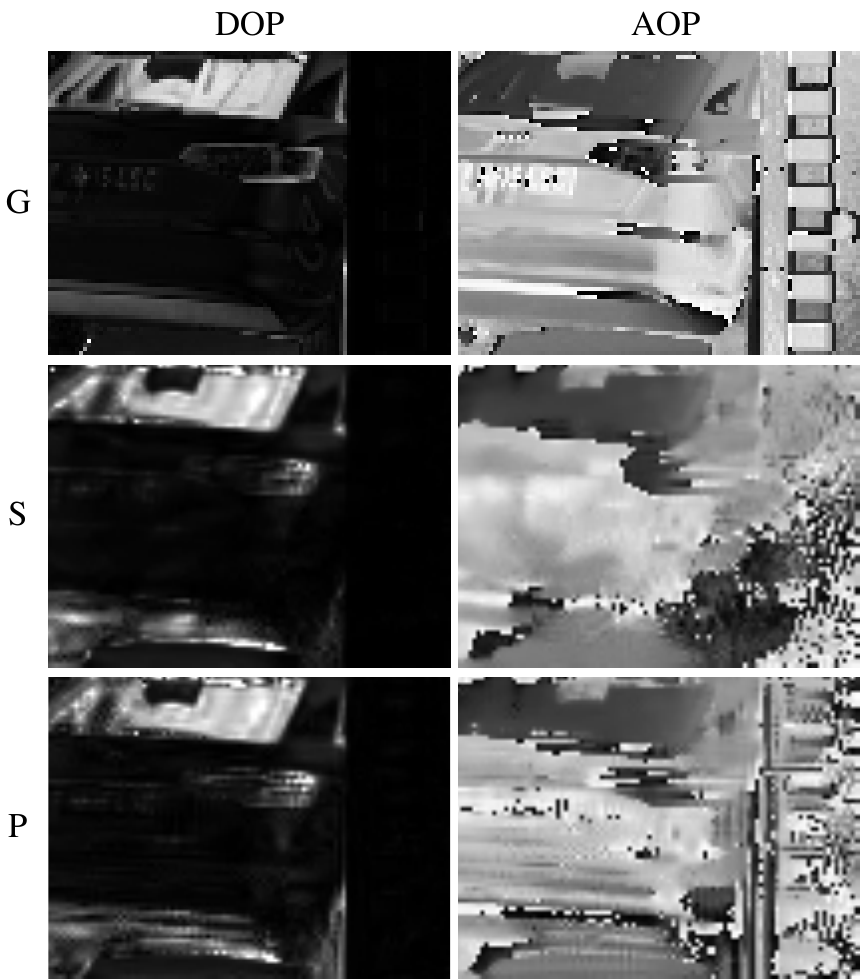}
\caption{Close up of `cars' image from figure~\ref{fig:simulatedImage4} (G: ground truth; S: BM3D Stokes; P: PBM3D). $\dop$ components are similar for the images denoised using BM3D Stokes and PBM3D, with slight differences noticeable in the car's bumper. The detail around the number plate of the car and panels on the right side of the image are more faithfully denoised using PBM3D than BM3D Stokes.}
\label{fig:carCrop}
\end{figure}

\begin{figure}
\includegraphics{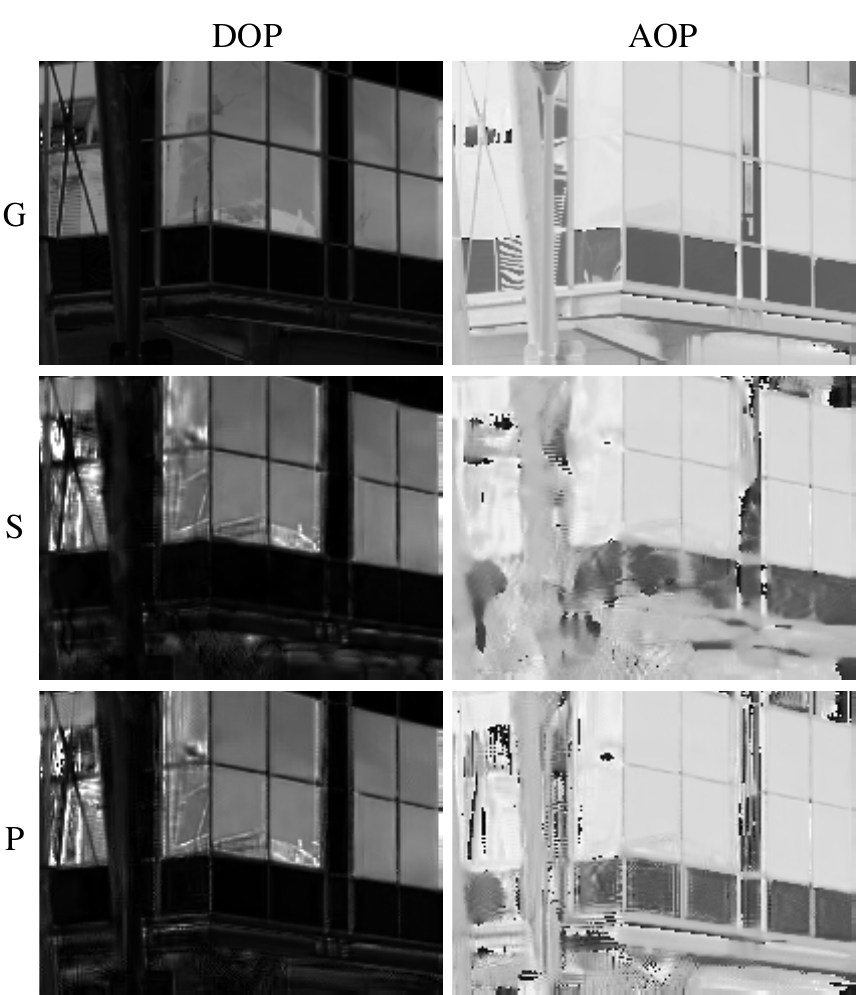}
\caption{Close up of `windows' image from figure~\ref{fig:simulatedImage8} (G: ground truth; S: BM3D Stokes; P: PBM3D). The $\dop$ component of the image denoised using PBM3D exhibits fewer artefacts than the imaged denoised using BM3D Stokes, especially underneath the window. In the $\aop$ components, the lower windows are much more faithfully represented by the image denoised using PBM3D than BM3D Stokes.}
\label{fig:windowCrop}
\end{figure}

Figures~\ref{fig:simulatedImage1},~\ref{fig:simulatedImage4}~\&~\ref{fig:simulatedImage8} show the denoised images corresponding to the $\sigma = 0.026$ row of table~\ref{tab:BasicPbm3dMeanTab}, as well as the ground truth and noisy images. It can be seen that as well as providing the greatest PSNR value, the visual quality of the images denoised using PBM3D is the greatest of the methods tested. In all three figures the $S_0$ component for the images denoised using BM3D, BM3D Stokes and PBM3D appear very similar to the ground truth, with the image denoised using Faisan appearing to be slightly less sharp. The $S_1$ and $S_2$ components of the images denoised using BM3D and Faisan appear to have more denoising artefacts than those denoised using BM3D Stokes and PBM3D. In the $\dop$ components the images denoised using PBM3D have cleaner edges, which are more similar to the ground truth than $\dop$ components denoised using all of the other methods, this is highlighted in figure~\ref{fig:windowCrop}, which shows a close up of the `window' images. The $\aop$ components denoised using PBM3D are notably more faithful to the original than the other methods, which can be seen in figures~\ref{fig:windowCrop}~\&{}~\ref{fig:carCrop}.

%

\subsection{Denoising real polarization imagery}

\begin{figure*}
\includegraphics{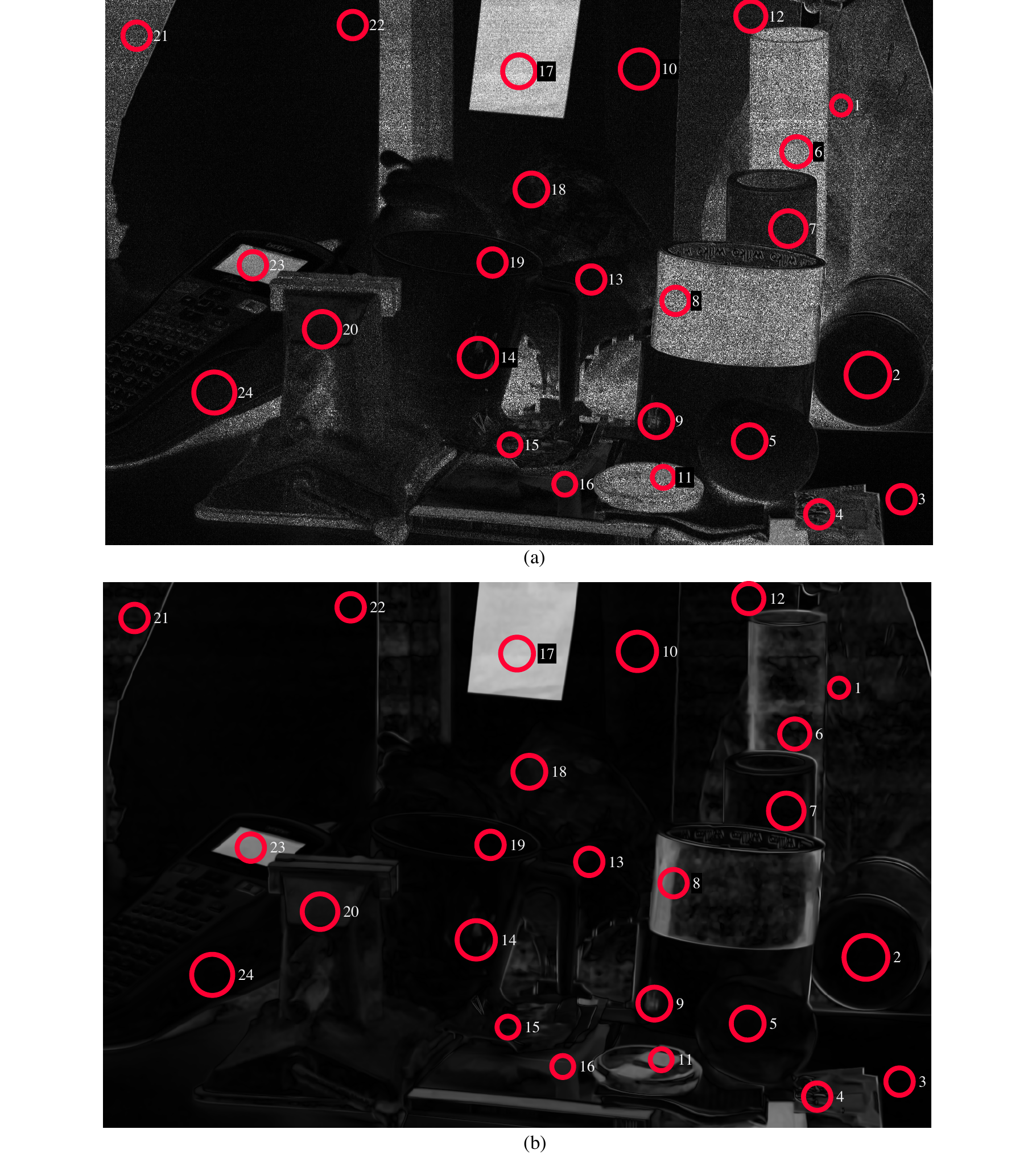}
\caption{$\dop$ image of a collection of lab objects, taken with an exposure of 0.0222s. (a): The image without denoising. (b): The image denoised using PBM3D. The circles indicate where the true $\dop$ value was measured using a spectrometer. It can be seen that the noisy image tends to show much larger $\dop$ values. The $\dop$ values measured at each point are shown in figure~\ref{fig:DopScatter}}
\label{fig:SpotNoisy}
\end{figure*}

To further test PBM3D with real, rather than simulated noise (as has been used so far), we used a DSLR camera with a rotatable polarizer to capture the three camera components, $I_0$, $I_{45}$, $I_{90}$, of a scene of several lab objects, using several exposure settings on the camera (table~\ref{tab:p}). The exposure setting was varied in order to vary the amount of noise present. The polarization images were then denoised using PBM3D. Figure~\ref{fig:SpotNoisy}(a) shows the $\dop$ of the captured image when the exposure was 0.0222s and figure~\ref{fig:SpotNoisy}(b) shows the $\dop$ of the same image, denoised using PBM3D. The effect of denoising is evident, with the perceptible noise in the noisy $\dop$ image being greater than for the denoised $\dop$ image.

\begin{figure}
\includegraphics{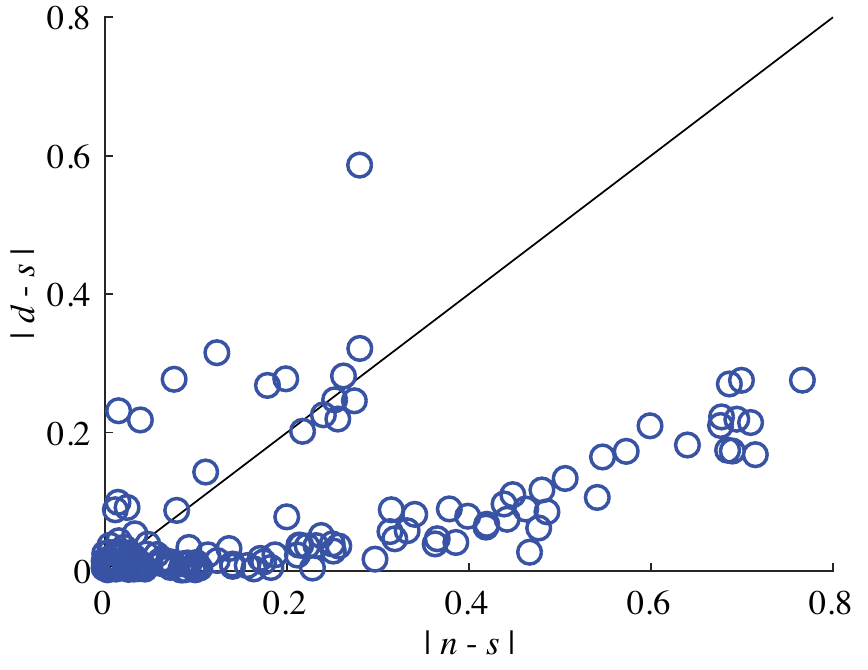}
\caption{The absolute value of the difference between the $\dop$ values measured using the spectrometer and the noisy images, $|n-s|$, and spectrometer and the denoised images, $|d-s|$. The noisy and denoised images had varying noise values, as given by table~\ref{tab:p}. The locations where the $\dop$ values were calculated are shown in figure~\ref{fig:SpotNoisy}. The black line indicates where $|d-s| = |n-s|$, it can be seen that using the denoised images for DoP measurements results in much greater agreement with measurements taken using the noisy image as for most measurements $|d-s| < |n-s|$. Table~\ref{tab:p} uses this data to show that denoising significantly reduces the error due to noise.}
\label{fig:DopScatter}
\end{figure}

In addition to the imaging polarimetry, we also measured the $\dop$ of several regions of the scene using a spectrometer. The intensity count was averaged across the wavelength range corresponding to the camera sensitivity (400-700nm) at three different orientations of a rotatable polarizer, $0^\circ$, $45^\circ$ and $90^\circ$. These mean intensities, $I_0$, $I_{45}$, $I_{90}$, were then used to calculate the $\dop$ using equation~\ref{eqn:dop}. The $\dop$ of the corresponding regions in the polarization images was also calculated using equation~\ref{eqn:dop} with a weighting on each of the camera components $(I_0, I_{45}, I_{90})$ to account for the separate $RGB$ channels, $I_i=0.299R + 0.587G + 0.11B$, which corresponds to the luminance, $Y$, of the $YUV$ colorspace. The absolute difference between the $\dop$ values from the spectrometry and from the imaging polarimetry with the noisy image and the same image denoised using PBM3D are shown in figure~\ref{fig:DopScatter}. The results were that the process of denoising extended the range of exposure time over which the imaging polarimetric values were the same as the spectrometry measurements. Table~\ref{tab:p} demonstrates that at an exposure time of $0.0222$s, when $\sigma \geq 0.0103$, the values of the $\dop$ from the noisy image become significantly different ($Wilcoxon, n = 24, V = 43, p = 0.002$) from those calculated using the spectrometry measurements. In contrast, when the images were denoised using PBM3D the exposure time could be bought down to 0.0056s ($\sigma \geq 0.0363$) before the $\dop$ values became different ($Wilcoxon, n = 24, V = 73, p = 0.040$). Therefore, denoising using PBM3D increases the accuracy of the measurements by reducing the effect of noise on the measurement, allowing approximately 3.5 times as much noise to be tolerated.

\begin{table}
\begin{tabular}{@{}cccccc@{}}
\toprule
& &\multicolumn{2}{c}{noisy} & \multicolumn{2}{c}{denoised} \\
\cmidrule(lr){3-4}
\cmidrule(lr){5-6}
& &\multicolumn{4}{c}{Wilcoxon $(n=24)$}\\
\cmidrule(lr){3-6}
exposure & estimated $\sigma$ & $V$ & $p$ & $V$ & $p$ \\
\midrule
0.1667 & 0.0021 & 154 & 0.8934 & 217 & 0.6575 \\
0.1000 & 0.0034 & 83 & 0.2700 & 191 & 0.6134 \\
0.0500 & 0.0055 & 74 & 0.0620 & 145 & 0.9589 \\
0.0222 & 0.0103 & 43 & \textbf{0.0024} & 157 & 0.9261 \\
0.0111 & 0.0199 & 18 & \textbf{0.0001} & 116 & 0.4154 \\
0.0056 & 0.0363 & 16 & \textbf{0.0000} & 73 & \textbf{0.0402} \\
0.0029 & 0.0678 & 7 & \textbf{0.0000} & 75 & \textbf{0.0022} \\
\bottomrule \end{tabular}
\caption{Estimated $\sigma$, the standard deviation of noise, and Wilcoxon test results for the data in figure~\ref{fig:DopScatter}. $\sigma$ values were estimated using the method in~\cite{faisan_joint_2012}. The Wilcoxon test indicates that when $\sigma \geq 0.0103$ the $\dop$ values calculated from the noisy image are significantly different to the $\dop$ values calculated from the spectrometer (bold indicates $p<0.05$). In contrast, the $\dop$ values calculated from the denoised images are significantly different when $\sigma \geq 0.0363$. Denoising therefore significantly reduces the effect of noise when $0.0103 \leq \sigma < 0.0363$.}
\label{tab:p}
\end{table}

\section{Conclusion}
\label{sec:conc}
Imaging polarimetry provides additional useful information from a natural scene compared to intensity-only imaging and it has been found to be useful in many diverse applications. Imaging polarimetry is particularly susceptible to image degradation due to noise. Our contribution is the introduction of a novel denoising algorithm, ‘PBM3D’, that when compared to state of the art polarization denoising algorithms, gives superior performance. When applied to a selection of noisy images, those denoised using PBM3D had a PSNR of 4.50dB greater on average than those denoised using the method of Faisan et al.~\cite{faisan_joint_2012}, and 0.84dB greater than those denoised using BM3D Stokes. PBM3D relies on a transformation from camera components into intensity-polarization components. We have given two algorithms for computing the optimal transformation matrix, and given the optimal for our system and dataset. We have also shown that if imaging polarimetry is used to provide $\dop$ point measurements, that denoising using PBM3D allows approximately 3.5 times as much noise to be present than without denoising for the image to still have accurate measurements.

%
%
%
%
%

\section{Funding Information}

EPSRC CDT in Communications (EP/I028153/1); Air Force Office of Scientific Research (FA8655-12-2112).

The authors thank C. Heinrich and J. Zallat for the use of their denoising code.

\bibliography{bibliography}

\ifthenelse{\equal{\journalref}{ol}}{%
\clearpage
\bibliographyfullrefs{sample}
}{}
 

\ifthenelse{\equal{\journalref}{aop}}{%
\section*{Author Biographies}
\begingroup
\setlength\intextsep{0pt}
\begin{minipage}[t][6.3cm][t]{1.0\textwidth} 
  \begin{wrapfigure}{L}{0.25\textwidth}
    \includegraphics[width=0.25\textwidth]{john_smith.eps}
  \end{wrapfigure}
  \noindent
  {\bfseries John Smith} received his BSc (Mathematics) in 2000 from The University of Maryland. His research interests include lasers and optics.
\end{minipage}
\begin{minipage}{1.0\textwidth}
  \begin{wrapfigure}{L}{0.25\textwidth}
    \includegraphics[width=0.25\textwidth]{alice_smith.eps}
  \end{wrapfigure}
  \noindent
  {\bfseries Alice Smith} also received her BSc (Mathematics) in 2000 from The University of Maryland. Her research interests also include lasers and optics.
\end{minipage}
\endgroup
}{}

\end{document}